\documentclass[conference]{IEEEtran}

\usepackage{longtable}% for long tables

\usepackage{times}
\usepackage{soul}
\usepackage{helvet}
\usepackage{courier}
\usepackage{booktabs}
\usepackage{multirow}
\usepackage[english]{babel}
\usepackage{mathrsfs}
\usepackage{graphicx}
\usepackage{amssymb}
\usepackage{url}
\usepackage{subfigure}
\usepackage{amsmath}
\usepackage{amsfonts}
\usepackage[ruled]{algorithm2e}
\usepackage{algorithmicx}
\usepackage{cite}
\usepackage{bbm}
\usepackage{threeparttable}
\usepackage{array}
\usepackage{arydshln}
\usepackage{color}

\newcommand{\mb}{\mathbf}

\newcommand{\rmsprop}{\textsc{RMSProp}}

\newcommand{\adagrad}{\text{AdaGrad}}

\newcommand{\adam}{\textsc{ADAM}}
\newcommand{\yadam}{\textsc{Gadam}}

\newcommand{\yb}{\textsc{BGadam}}
\newcommand{\bos}{\text{boosting}}
\newcommand{\ga}{\text{genetic algorithm}}

\usepackage{longtable}% for long tables

\usepackage{booktabs}
\title{BGADAM: Boosting based Genetic-Evolutionary ADAM for Neural Network Optimization}

\author{Jiyang Bai, Yuxiang Ren, Jiawei Zhang\\
	Florida State University\\%, IFM Lab\\
	bai@cs.fsu.edu, yuxiang@ifmlab.org, jiawei@ifmlab.org
}

\begin{document}

\maketitle

\begin{abstract}
For various optimization methods, gradient descent-based algorithms can achieve outstanding performance and have been widely used in various tasks. Among those commonly used algorithms, {\adam} owns many advantages such as fast convergence with both the momentum term and the adaptive learning rate. However, since the loss functions of most deep neural networks are non-convex, {\adam} also shares the drawback of getting stuck in local optima easily. To resolve such a problem, the idea of combining {\ga} with base learners is introduced to rediscover the best solutions. Nonetheless, from our analysis, the idea of combining {\ga} with a batch of base learners still has its shortcomings. The effectiveness of {\ga} can hardly be guaranteed if the unit models converge to close or the same solutions. To resolve this problem and further maximize the advantages of {\ga} with base learners, we propose to implement the {\bos} strategy for input model training, which can subsequently improve the effectiveness of {\ga}. In this paper, we introduce a novel optimization algorithm, namely \textbf{B}oosting based \textbf{G}enetic \textbf{{\adam}} ({\yb}). With both theoretic analysis and empirical experiments, we will show that adding the {\bos} strategy into the {\yb} model can help models jump out the local optima and converge to better solutions.
\end{abstract}
%\begin{keywords}
%Optimization Algorithm; Genetic Algorithm; Boosting.
%\end{keywords}

%------------------------------------------
\section{INTRODUCTION}\label{sec:intro}

Deep learning models have achieved impressive success in many areas, which include the computer vision~\cite{imagenet}, natural language processing~\cite{rnn,lstm} and recently appeared graph neural networks~\cite{gnn, gnn_2}. By owning a large number of variables and non-linear functions (e.g., the Relu function~\cite{relu}), deep learning models can fit extremely complex data and learn the hidden representations of them. To let the deep learning models achieve outstanding performance, one of the most important procedures is training the models with appropriate optimization algorithms. During the training process of deep learning models, the essence is to search for the global optima of the loss functions. This process can be represented by
\begin{equation}
\min_{\mb{w} \in \mathcal{W}} f(\mb{X},\mb{y}; \mb{w})
\end{equation}
Here, $f(\cdot ,\cdot ;\mb{w})$ denotes the loss function with variables $\mb{w}\in \mathcal{W}$, where $\mathcal{W}$ is the solution space of variables; $\mb{X}$ and $\mb{y}$ denote the features and labels of the training data respectively. Up to now, the most commonly used optimization algorithms are based on the gradient descent, for example the stochastic gradient descent (SGD), SGD with momentum~\cite{sgd_overview}, {\adagrad}~\cite{adagrad}, {\rmsprop}~\cite{rmsprop} and {\adam}~\cite{ADAM}. Among these algorithms, {\adam} is the most widely used one and has been proven to be powerful in various tasks thanks to its fast convergence rate. However, {\adam} still has the common drawback of easily converging to the local optima and getting stuck into it, especially when dealing with the non-convex problems~\cite{understand_difficulty_tra_nn}. To solve the problem of converging to the local optima, the idea of integrating {\ga} and multiple learners have been proposed such as {\yadam}~\cite{yadam}, which can combine the advantages of both the {\ga} and the {\adam}. During the training process, initially multiple input models are trained simultaneously. From these input models, a batch of (parent) model pairs are selected, and {\ga} is implemented to the variables of these parent models after they converge to the local optima. Applying the {\ga} can help the trained parent models ``jump out'' of local optimal points when dealing with the non-convex problem. The ``jump out'' operation is implemented through reorganizing the variables of the parent models. Even though {\ga} with multiple learners has the potential capability of approaching more optimized solutions, the actual effect still can not be guaranteed. If most of the input models, or parent models converge to the close or even the same local optima, the {\ga} operation might become meaningless since the output models of the variables reorganizing still locate at the local optima. We denote this phenomenon of failing to jump out of local optima as the ``local sticking''. To avoid the ``local sticking'' problems and increase the capability of converging to better solutions, one concrete idea is to let the parent models converge to divergent local optima. In this way, the {\ga} is more likely to generate models with variables at different locations and subsequently increase the possibility of converging to better solutions and achieving better performance.

In this paper, we will propose a new optimization algorithm, namely \textbf{B}oosting based \textbf{G}enetic \textbf{ADAM} ({\yb}) to solve the ``local sticking'' problem mentioned above and further achieve better performance compared with {\yadam} algorithm. With the support of the boosting strategy, {\yb} can improve the performance of the optimization by differentiating the parent models, and meanwhile guarantee the convergence of the algorithm theoretically. Boosting strategy~\cite{boosting, boosting2} utilizes the interactions among base learners to accomplish a more effective model training. The interaction is achieved by redistributing the dataset to modify the training set for each input model. In each training iteration, the {\bos} strategy samples training data from the training dataset with replacement subject to different weights. After the training process of one input model, the weights of all the data samples will be updated according to the classification results of this model. Based on the updated weights, the training set will be sampled under the updated distribution, and newly sampled training data will be assigned to the next input model. Meanwhile, this newly sampled dataset inclines to contain more data misclassified by the previous input model. In this way, by proposing the boosting strategy, different parent models will be trained with different training examples. In other words, the parent models trained by different training set will subsequently converge to divergent local optima with a higher probability. 
Therefore, {\yb} can effectively diversify the learned parent models, which can further improve the learning performance.

The following part of the paper is organized as follows. In Section~\ref{sec:relatedwork}, we will talk about some related works. In Section~\ref{sec:method}, we will cover more details about our proposed {\yb} algorithm, whose effectiveness will be analyzed in Section~\ref{sec:theorecital_analysis}, and performance will be tested with extensive experiments in Section~\ref{sec:experiment}. Finally, we will conclude this paper in Section~\ref{sec:conclusion}.

%------------------------------------------
%-----------------------------------------------------------------
\begin{table}
	\vspace{-10pt}
	\centering
	\caption{Notations and Terminologies Definitions}
	\begin{tabular}{|c|c|}
		\hline
		Notation&Definition\\
		\hline
		$m$&Number of samples in training set\\
		$\mathcal{D}_j$&Training set for $j_{th}$ model\\
		$\mb{z}$&Weight vector of training samples\\
		$M$&Input model\\
		$\bar{M}$&Trained input model\\
		$\mb{\bar{w}}_i$&Variables of model $\bar{M}_i$\\
		$C$&Child model\\
		$\mb{w}_i$&Variables of child model $C_i$\\
		$N$&New generation model\\ 
		$\epsilon$&Training error rate\\
		$g$&Number of input models in each generation\\
		$K$&Number of generations\\
		$G^{(k)}$&$k_{th}$ generation\\ 
		\hline
	\end{tabular}
	\label{tab:notation}
	%	\vspace{-10pt}
\end{table}
%---------------------------------------------------------------
\section{RELATED WORKS} \label{sec:relatedwork}

\noindent\textbf{Deep Learning Models}: Deep learning models have achieved state-of-the-art results in recent years, whose representative examples include feed-forward deep neural networks (DNN)~\cite{dbm, autoencoder},  convolutional neural networks (CNN)~\cite{imagenet}~\cite{GBDR}, and recurrent neural networks (RNN)~\cite{rnn,lstm}. DNN models mainly consist of several layers of the linear transforms and the non-linear activation functions. The combination of these linear and non-linear transforms between input features and outputs enables the DNN to fit more complex patterns hidden in the training data. Inspired by the great success of DNN models, more attention had been focused on transplanting the deep model ideas onto other types of data such as the image type and the sequential type data. To deal with the image data, CNN has been proposed and already shown outstanding performance on various computer vision tasks; for the sequential data, the RNN model came up. RNN~\cite{critical_review_rnn, newtype_speechrecongnition} models are connectionist models with the ability to pass information across sequence elements while conducting sequential data one element at a time. Besides, there also exist many other types of deep learning models to tackle different types of data, e.g., the graph neural networks~\cite{gnn,gnn_2, huang2018adaptive, hamilton2017inductive}, graph convolutional networks~\cite{gcn} and graph attention networks~\cite{gat} to deal with graph type data; the Generative Adversarial Networks (GAN)~\cite{gan} to train discriminative models and generative models, etc.

%-------------------------------------------------------
\begin{figure*}[t]
	%\subfigure[]{ 
	%\begin{minipage}[l]{1.2\columnwidth}
	\centering
	\includegraphics[width=1\textwidth, height=0.25\textheight]{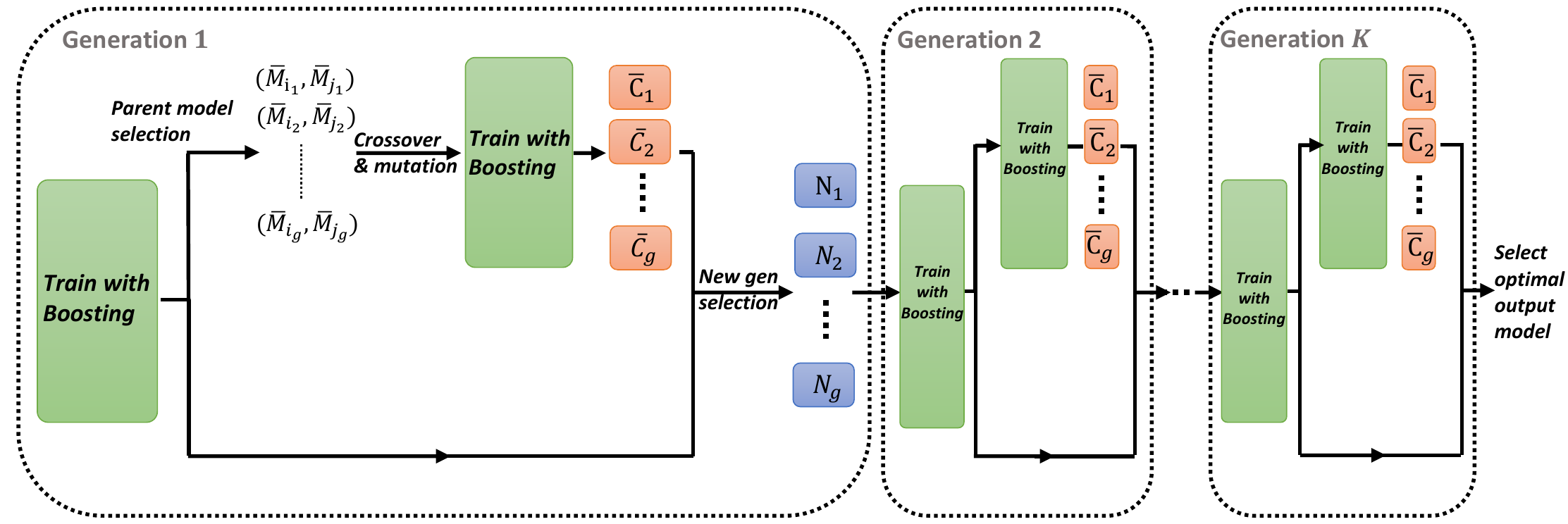}
	%\end{minipage}
	%	}
	\vspace{-15pt}
	\caption{{\yb} structure}%\label{fig:y+b}
	\label{fig:yb}
	\vspace{-15pt}
\end{figure*}
%--------------------------------------------------------

\noindent\textbf{Genetic Algorithm}: Genetic algorithm~\cite{ga, whitley1994genetic, loshchilov2016cma, jaderberg2017population,cui2018evolutionary} is a family of computational algorithms inspired by creature evolution in the natural environment. It encodes a potential solution to a specific problem on chromosome-like data structures and applies recombination (reorganization) operators to these structures~\cite{ga_tutorial}. This process is also called crossover~\cite{ga_convergence}. Besides, {\ga} also adds random mutations on the solutions to mimic the natural world's real gene mutation process. In this way, for some models stuck in local optimal points, {\ga} can help them jump out by switching specific variables. During the generating process, the above crossover and mutation operations are implemented to a bunch of models, which are called the population set. After that, the models with the worst performance (the fitness score) will be eliminated. Finally, individuals with relatively better performance are retained.

\noindent\textbf{Optimization Algorithms}: The main purpose of the optimization algorithms is to closely approach the best solution of the loss functions during the backpropagation. So far most commonly used optimizers are based on the gradient descent~\cite{sgd_overview}, such as the stochastic gradient descent (SGD), the {\adagrad}~\cite{adagrad}, the {\rmsprop}~\cite{rmsprop} and the {\adam}~\cite{ADAM}. 
SGD performs a variable update for each training example $\mb{X}[i,:]$ and label $\mb{y}[i]$, it can be expressed as:
\begin{equation}
\mb{w} = \mb{w} - \eta\cdot\nabla_{\mb{w}} f(\mb{X}[i,:], \mb{y}[i])
\end{equation}
where $\eta$ is the learning rate and $\nabla_{\mb{w}}$ is the derivative of the loss function regarding variable $\mb{w}$. The advantages of SGD include fast speed and getting rid of redundancy~\cite{sgd_overview}. SGD with momentum~\cite{moment, impor_mom_deeplearning} is a method that helps accelerate SGD in the relevant directions. 
SGD with momentum updates variables with the following equations:
\begin{equation}
\begin{split}
&\mb{v}_t =  \gamma\cdot \mb{v}_{t-1} + \eta\cdot \nabla_{\mb{w}} f(\mb{w})\\
&\mb{w}_t = \mb{w}_t -\mb{v}_t
\end{split}
\end{equation}
where $\gamma$ is the momentum term weight. The momentum term accelerates variable updates for dimensions whose gradients are in the same direction as historical gradients and decelerates updates for dimensions whose gradients are the reverse. Therefore its convergence process will be faster. However, the unified learning rate for all variables can lead to some problems. If variables have different scales, we should not update them with the same rate. Some variant algorithms have been proposed to solve this problem, such as the {\adagrad} and the {\rmsprop}. {\adagrad}~\cite{adagrad} adapts the learning rate to the parameters, performing larger updates for infrequent and smaller updates for frequent parameters. {\rmsprop}~\cite{rmsprop} is a refined version of the {\adagrad} that applies the moving average of the historical gradients to adapt the learning rates. {\adam}~\cite{ADAM} is proposed based on the SGD and the momentum, which computes individual adaptive learning rates for different variables. Similar to the SGD with momentum and the {\rmsprop}, {\adam} records the first-order momentum $\mb{m}_t$ and the second-order momentum $\mb{v}_t$ of the gradients and computes the bias-corrected version of them respectively.

\noindent\textbf{Ensemble Learning}: Ensemble learning~\cite{boosting2, dietterich2002ensemble, zhang2012ensemble} tries to train multiple learners to solve the same problem. In contrast to ordinary learning approaches that attempt to construct one model from the training data, ensemble methods try to construct a set of learners and combine them. The commonly used ensemble methods include boosting and bagging. Boosting~\cite{boosting,boosting2} generally creates a batch of weak learners, and each learner is trained based on the sampled training data under the distribution decided by the previous learner. In other words, each learner can focus more on the data samples being misclassified by the previous learner. Bagging adopts the bootstrap distribution for generating different base learners. It applies bootstrap sampling~\cite{bootstrap} to obtain the data subsets for training the base learners.

%Given a training dataset containing $m$ training examples,firstly selects  a batch of $m$ training examples by random sampling with replacement. Then $g$ training sets from the same original training data by applying the {\bos} sampling $g$ times~\cite{boosting2}, which will be used to train $g$ different base models respectively in boosting.

%------------------------------------------
\section{{\yb} LEARNING ARCHITECTURE}\label{sec:method}

\begin{figure}[t]
	\vspace{-20pt}
	\centering
	\includegraphics[width=0.4\textwidth,height=0.25\textheight]{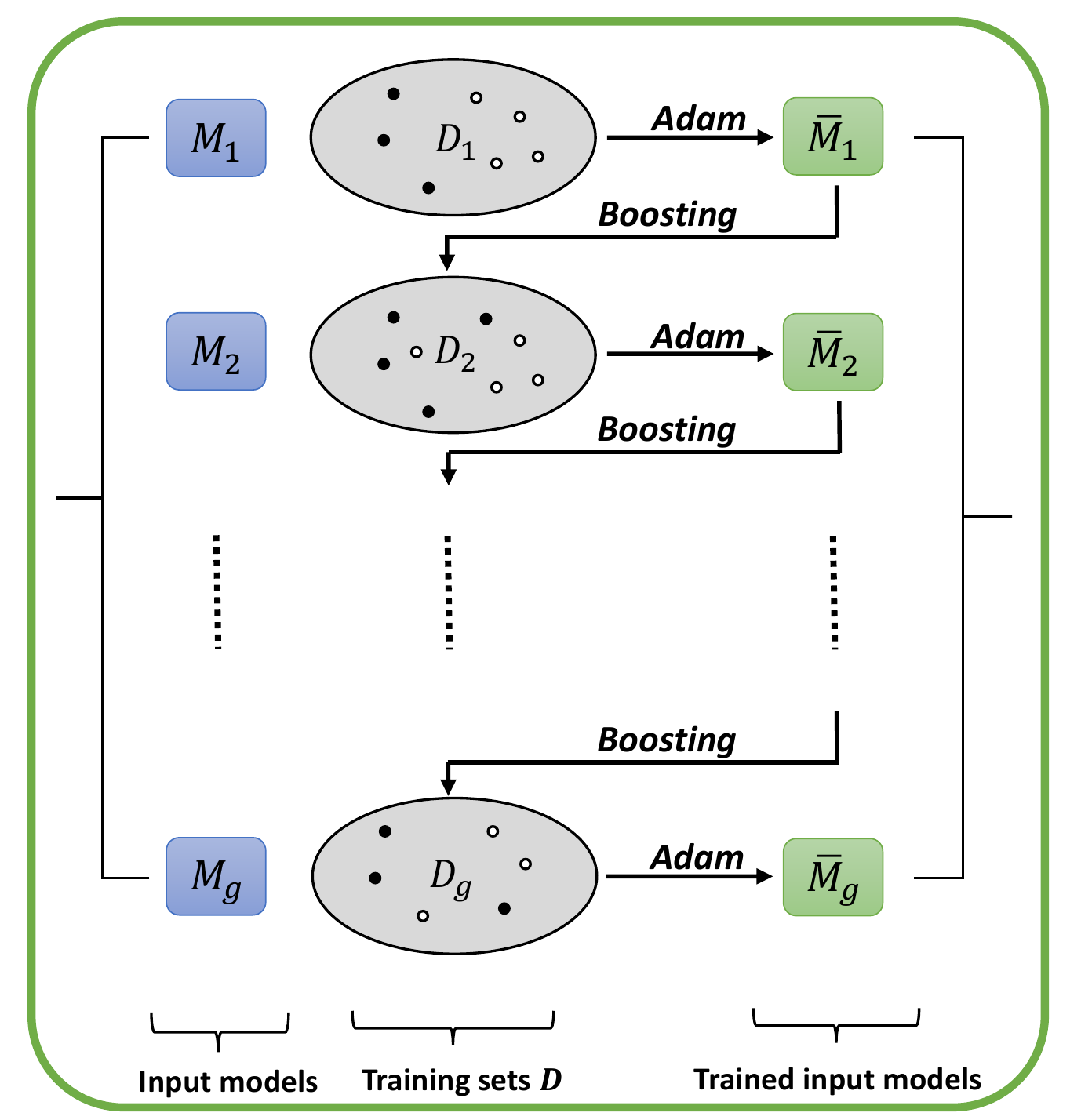}
	\caption{Boosting based {\adam}}%\label{fig:y+b}
	\label{fig:bos}
	\vspace{-10pt}
\end{figure}

In this section, we will introduce more details about the {\yb} optimizer. The architecture of {\yb} is illustrated in Figure~\ref{fig:yb}. In the architecture, the green frame denotes the {\bos} strategy based {\adam}, whose detailed structure is provided in Figure~\ref{fig:bos}. Next, we will provide more detailed descriptions of both the architecture and the involved learning components. The notations and terminologies we have employed in this paper are presented in Table~\ref{tab:notation}.

{\yb} adopts the {\bos} strategy to increase the diversity of the input models, which can further resolve the ``local sticking'' problem of the parent models. The learning process of {\yb} involves multiple generations, which can be denoted as $G^{(1)}, G^{(2)},\cdots, G^{(K)}$ respectively ($K$ is the total generation count). In each generation, $g$ input models will be trained along with the {\bos} based {\adam} at first; then a batch of trained input models will be selected to compose the parent model pairs; after that, the child models will evolve according to the {\ga}; finally, among both the input models and the generated child models, the same number (e.g., $g$) of new generation models will be selected as the output models of the current generation and as the input models of the next generation. After a predefined number of iterations (or some converging criteria has been satisfied), the training will stop, and the best new-generation models will become the output model. We will refer to each procedure in the following parts.
\vspace{-5pt}

\subsection{Training Input Models with Boosting Strategy}
In each generation $G^{(k)}, k \in \{1,2,\dots ,K\}$,  $g$ input models $\{M_1, M_2, \dots, M_g\}$ are trained with {\bos} based {\adam}. More detailed information about the {\bos} based {\adam} training will be covered in Section~\ref{subsec: bos_training}. The essence of {\bos} strategy is to train multiple models with collaborating policy~\cite{boosting2}, then combine these models to induce the final results. For each model, the collaborating is achieved by sampling different sub training sets according to the previous model's performance. Specifically, during the training process of $g$ input models, the same size $g$ training sets $\{\mathcal{D}_1, \mathcal{D}_2, \dots, \mathcal{D}_g \}$ will be generated. For each training set $\mathcal{D}_j$, its data samples are sampled from the entire training set, under the distribution determined by the performance of model $M_{j-1}$. For our classification task, the data samples having been misclassified by $M_{j-1}$ will be assigned higher probabilities to be sampled in $\mathcal{D}_j$. In this way, {\bos} strategy allocates each input model with different training data. Since the different training sets will allow the input model variables to be updated by {\adam} differently, these $g$ input models are more likely to converge to different local optimal points. Such a characteristic gives the following {\ga} more advantages, and we will discuss this in the later parts. Moreover, {\bos} strategy not only allows each input model to learn information hidden in data, but also refers to previously trained models. We also call this process the interaction among the input models. 
After the training process of $g$ input models, $\{\bar{M}_1, \bar{M}_2,\dots, \bar{M}_g\}$ refers to the trained input models, and then parent model pairs $\{(\bar{M}_{i_1}, \bar{M}_{j_1}),\dots, (\bar{M}_{i_g}, \bar{M}_{j_g})\}, i, j \in \{1,2,\dots ,g\}$ will be chosen from them based on special mechanism. Next, the child models will evolve from them by the {\ga}, which essentially helps the parent models jump out of the local minimum and discover global optima to achieve better learning performance.
%-------------------------------------------------------
%--------------------------------------------------------

\subsection{Parent Model Selection and Fitness Evaluation}

Similar to the evolutionary laws in the natural world, where individuals with better performance own higher possibilities to become the parents, our proposed {\yb} also inclines to choose input models with better performance to participate in the evolving process. Before the evolving operations, from the input models trained by {\bos} based {\adam}, a batch of parent model pairs will be selected according to the fitness evaluation. The fitness evaluation aims to evaluate each trained input model by computing their fitness scores for the learning setting, and a better fitness score means better performance of the target model. Correspondingly, input models with better fitness scores would be more likely to be selected as the parent models lately. Thus, the function of fitness evaluation is akin to evolutionary laws. 

In the $k_{th}$ generation $G^{(k)}$, for each trained input model $\bar{M}_i$, we calculate its loss on validation set $\mathcal{V}$ (sampled from the training set) as the fitness score, which can be denoted as follows:
\begin{equation}
\mb{l}[i] = \sum_{(\mb{X}[j,:], \mb{y}[j]) \in \mathcal{V}} l(\mb{X}[j,:],\mb{y}[j]; \mb{w}_i)
\end{equation}
where $l(\cdot,\cdot;)$ is the cross-entropy loss function, $\mb{w}_i$ denotes the variables of model $\bar{M}_i$, and vector $\mb{l} \in \mathbb{R}^{g}$ contains the computed fitness scores of all these $g$ unit models in the current generation. By calculating the loss of input models on the validation set, we can judge these input models: the less the loss is, the better performance model will have in general. However, directly using the loss terms for parent model selection may not work well because the range of $\mb{l}[i]$ might vary in a large scale. So for each $\mb{l}[i]$, we utilize the normalized $\hat{\mb{l}}[i] = \frac{\mb{l}[i] - min(\mb{l})}{max(\mb{l}) - min(\mb{l})}$ to calculate the selecting probability for input model $\bar{M}_i$ to become the parent model as follows: 
\begin{equation}
\mb{p}[i] = \frac{\exp(-\hat{\mb{l}}[i])}{\sum_{j=1}^{g} \exp(-\hat{\mb{l}}[j])}
\label{equ:probability}
\end{equation}
where $\mb{P} \in \mathbb{R}^{g}$. According to the probability vector $\mb{p}$, $g$ different model pairs $\{(\bar{M}_{i_1}, \bar{M}_{j_1}),\dots, (\bar{M}_{i_g}, \bar{M}_{j_g})\}$ will be sampled as parent models. Specifically, for each model in these pairs, it is probability sampled by $\mb{p}$ with replacement.

\subsection{Crossover and Mutation}\label{subsec:crossover_mutation}

The genetic laws of individuals' genes are common in the natural world to obtain better offspring. 
Inspired by the recombination and mutation rules of genes in genetic laws, the {\ga} also carries out similar operations, named crossover and mutation separately. The {\ga} includes crossover and mutation operations for creating child models generation, where the model variables are treated as the chromosome of gene, respectively. 
\begin{itemize}
	\item \textbf{Crossover}: Given a parent model pair $(\bar{M}_{i_n}, \bar{M}_{j_n})$ with respective variable vectors $\mb{\bar{w}}_{i_n}$ and  $\mb{\bar{w}}_{j_n}$, crossover generates their child model with variable vector $\mb{w}_n$, whose entry $\mb{w}_n[h]$ can be represented as
	\begin{equation}
	\begin{split}
	\mb{w}_n[h] &= \mathbbm{1}(\textbf{rand}\leq threshold)\cdot \mb{\bar{w}}_{i_n}[h] \\
	&+ \mathbbm{1}(\textbf{rand} > threshold)\cdot \mb{\bar{w}}_{j_n}[h]
	\end{split}
	\label{equ:crossover}
	\end{equation}
	In the Equation~\ref{equ:crossover}, $\mathbbm{1}(\cdot)$ is the indicator function, $\textbf{rand}$ is a random number uniformly distributed in $[0,1]$, $threshold = \frac{\mb{p}[i_n]}{\mb{p}[i_n] + \mb{p}[j_n]}$ and $\mb{\bar{w}}_{i_n}[h]$, $\mb{\bar{w}}_{j_n}[h]$ denotes the $h_{th}$ variable of the parent models $\bar{M}_{i_n}$ and $\bar{M}_{j_n}$ respectively. By applying the evolving rule in Equation~\ref{equ:crossover} to each parent model pairs, the model with relatively better performance would retain more its own variables. 
	\item \textbf{Mutation}: Similar to crossover, the mutation can be expressed as
	\begin{equation}
	\begin{split}
	\mb{w}_n[h] &= \mathbbm{1}(\textbf{rand} \leq p_m)\cdot \mathcal{N}(0, 0.01)\\
	&+ \mathbbm{1}(\textbf{rand} > p_m)\cdot \mb{w}_n[h]
	\end{split}
	\label{equ:mutation}
	\end{equation}
	where $p_m = p\cdot(1 - \mb{p}[i_n] - \mb{p}[j_n])$ and $p$ is a pre-defined value (e.g., $p = 0.01$), $\mathcal{N}(0, 0.01)$ is a random number sampled from the normal distribution with 0 mean and 0.01 variance. Equation~\ref{equ:mutation} can make sure the child models of parents with higher $\mb{p}$ values to be less likely to mutate. In other words, the mutaton mechanism in Equation~\ref{equ:mutation} inclines to preserve the variables inherited from parents that having better performance.  
\end{itemize}
Formally, via crossover and mutation, we can represent the generated $g$ child models as $\{C_1, C_2, \cdots, C_g\}$, which will be further trained via the {\bos} based {\adam}. Those trained child models can be denoted as $\{\bar{C}_1, \bar{C}_2, \cdots, \bar{C}_g\}$ respectively.

%----------------------------------

\begin{algorithm}[tb]
	\caption{{\yb}} %算法的名字
	\KwIn{ %算法的输入， \hspace*{0.02in}用来控制位置，同时利用 \\ 进行换行
		Input models $\{M_1, M_2, \dots,M_g\}$; training feature $\mb{X}$; training label $\mb{y}$}
	\KwOut{Final model}
	%	\begin{algorithmic}[1]
	\For {$i = 1,2,\dots , K$}
	{
		$\mb{z} = [\frac{1}{m},\frac{1}{m},\dots,\frac{1}{m}$] and $\mathcal{D}_1 = (\mb{X}, \mb{y})$; \\
		$\mb{l} = []$ \enspace; /* To record loss. */ \\
		\For {$j = 1,2,\dots, g$}
		{
			Train $M_j$ as $\bar{M}_j$ using $\mathcal{D}_j$ dataset;\\
			$\mb{l}[j] =$ The loss of $\bar{M}_j$ on $\mathcal{V}$;\\
			Update $\mb{z}$ by Equation (\ref{equ:W1}) and Equation (\ref{equ:sum});\\
			Produce $\mathcal{D}_{j+1}$ by sampling subject to $\mb{z}$;\\
			Compute $\mb{p}[j]$ according to Equation (\ref{equ:probability});\\
		}
		\For {$h = 1,2,\dots,g$}
		{
			Select $\bar{M}_{i_k}$ and $\bar{M}_{j_k}$ according to $\mb{p}$;\\
			Generate $C_k$ by Equations (\ref{equ:crossover}) and (\ref{equ:mutation});\\
			Train $C_h$ as $\bar{C}_h$ using $\mathcal{D}_h$;\\
			$\mb{l}[h] =$ The loss of $\bar{C}_h$ on $\mathcal{V}$;\\
		}
		\For {$l = 1,2,\dots,g$}
		{
			Select model with smallest value in $\mb{l}$ as $N_l$;\\
			Delete the smallest value in $\mb{l}$;\\
		}
	}
	
	\Return $N_1$ model
	%	\end{algorithmic}
	\label{alg:gadam}
\end{algorithm}

%--------------------------------------
\subsection{New Generation Selection}
The selection of the new generation of models also follows the evolutionary law: the best and strongest individuals will survive. The new generation models are selected among both the trained child models and input models. In this way, it is certain that the selected new generation of models will not deteriorate compared with the input models. This idea also conforms to the rule of survival of the fittest in natural selection. For each model in $\{\bar{M}_1,\bar{M}_2,\dots ,\bar{M}_g\}\bigcup \{\bar{C}_1,\bar{C}_2,\dots ,\bar{C}_g\}$, its loss on the validation set will be recorded. Finally, we will select the top $g$ models with the smallest losses as the new generation models, which can be denoted as $\{N_1,N_2,\dots ,N_g\}$ and they will also serve as the input models for the next generation. Such an iterative model learning and the evolving process will continue until finally converging, and the optimal output model (with the lowest loss on validation set) in the last generation will be selected as the final output model. We will also give the convergence analysis of {\yb} in Section~\ref{subsec:convergence_ana} and briefly mention the convergence results in Section~\ref{subsec: alg_convergence}.

%While training $M_i$, we also record its performance on validation set. In this paper, our validation set is selected as part of the training set. By evaluating their performance on validation set we can choose parent pairs. Details are shown in Algorithm~\ref{alg:pair}.

\subsection{Boosting Strategy based {\adam}}\label{subsec: bos_training}
Boosting refers to a family of algorithms that can convert weak learners to strong learners. The core idea of {\bos} is to correct the mistakes made by previous learners (models) and let the current model focus more on the data examples being misclassified by prior models. So during the training process of the current model, the training set will be different from those for prior models. In this section, we will talk more about the {\bos} based {\adam} learning algorithm adopted in {\yb}.

\subsubsection{Motivation}

\begin{figure}[t]
	\vspace{-25pt}
	\centering
	\includegraphics[width=0.38\textwidth]{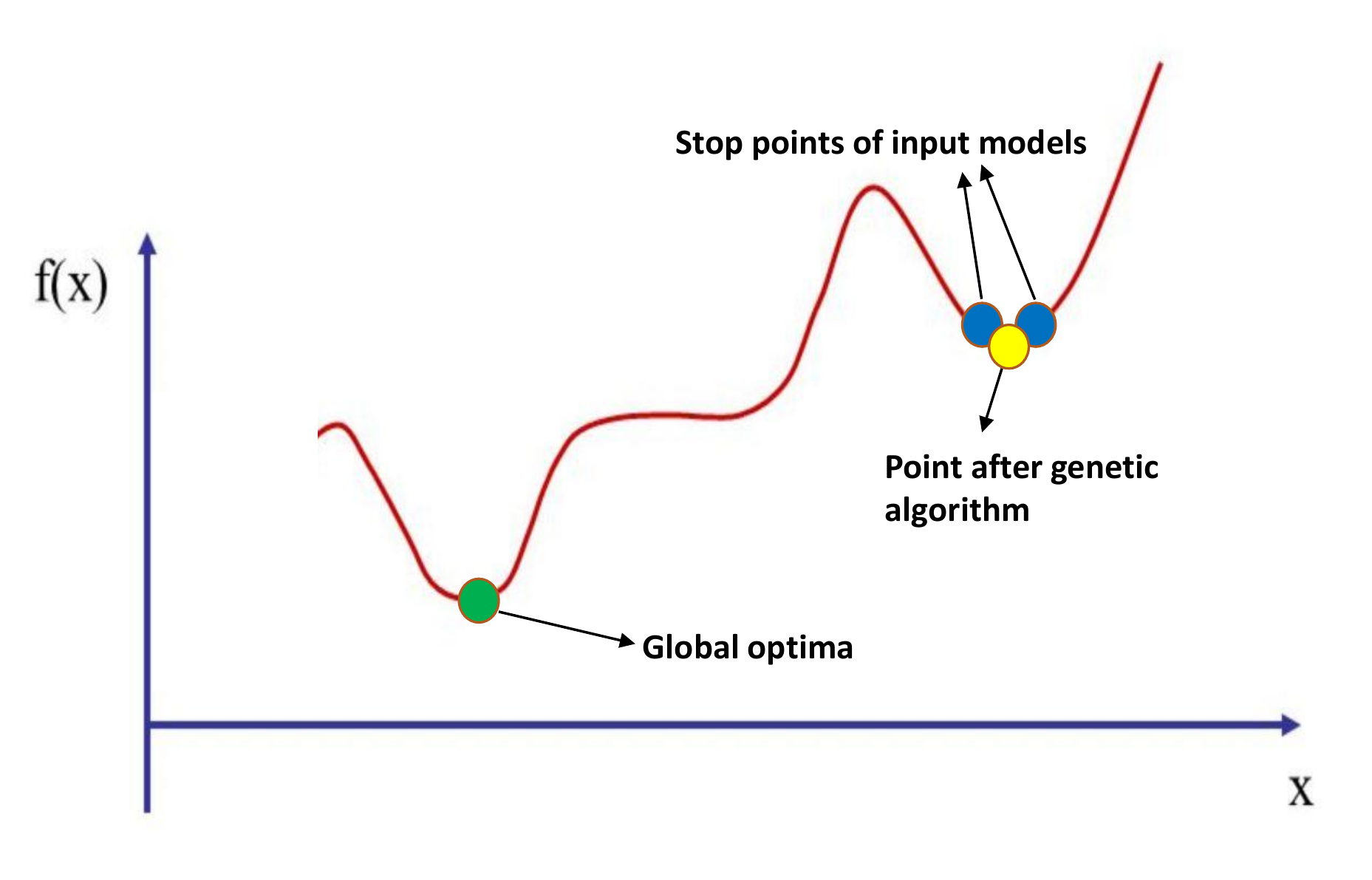}
	\vspace{-10pt}
	\caption{Local sticking situation}
	\label{fig:result_4}
	\vspace{-10pt}
\end{figure}

Prior to adding the {\bos} strategy, {\ga} is only combined with {\adam} to rediscover the best solutions of the models. 
%It not only provides the opportunities to search for solutions from multiple points, but also allows models to jump out of local optima and reach potential global optimal points. 
However, we observe that {\ga} cannot really resolve such a problem completely. For instance, assume that after training by {\adam}, all input models converge to the same or close local optima of the loss function, which can be eshibited by Figure~\ref{fig:result_4}. Under such circumstances, the {\ga} (crossover and mutation) can hardly help child models jump out of local optima because the parent models are too similar on the learned model variables. Assume the variables of $\bar{M}_{i_n}$ has $\mb{\bar{w}}_{i_n} \in U(\mb{o}, \delta) = \{\mb{w} |\left\| \mb{w}-\mb{o}\right\|_2 \leq \delta \}, \forall i_n \in \{1,2,\dots, g\}$, where $\mb{o}$ represents a local optima and $\delta$ is a default small value. Here, $U(\mb{o}, \delta)$ represents a neighborhood region around $\mb{o}$ in the variable space. Then for parent model pairs $\bar{M}_{i_n}$ and $\bar{M}_{j_n}$ located in $U(\mb{o},\delta)$, we can compute the variables of their child model $C_n$ as $\mb{w}_n[h] = \beta \cdot \mb{\bar{w}}_{i_n}[h] + (1-\beta)\cdot \mb{\bar{w}}_{j_n}[h]$, where $\beta \in \{0,1\}$, which is exactly the crossover operation in Section~\ref{subsec:crossover_mutation}. Then we have
\begin{equation}
\begin{split}
\left\| \mb{w}_n - \mb{o} \right\|_2 &\leq \left\| \mb{w}_n - \mb{\bar{w}}_{i_n}\right\|_2 + \left\|\mb{\bar{w}}_{i_n} - \mb{o}\right\|_2 \\
&\leq \left\| \mb{w}_n - \mb{\bar{w}}_{i_n} \right\|_2 + \delta\\
&\leq  \left\| \mb{\bar{w}}_{i_n} - \mb{\bar{w}}_{j_n} \right\|_2 + \delta \\
&\leq \left\|\mb{\bar{w}}_{i_n} - \mb{o} \right\|_2  + \left\| \mb{\bar{w}}_{j_n} - \mb{o} \right\|_2 + \delta \\
&\leq 3\delta
\end{split}
\end{equation}
and observe that after the crossover, $C_n$ still locates in the neighborhood region of the local optima $\mb{o}$. 
%Even after crossover and mutation, child models' variables(or parameters) have high similarity with their parents. 
This phenomenon is exactly the ``local sticking" situation we have mentioned in the Introduction section. It is true that mutation operation can assist in jumping out of the local optima. However, according to Equation (\ref{equ:mutation}) the parent models with a relatively small loss on the validation set will lead to a much lower mutation rate $p_m$ for their child models. Thus mutation cannot solve the local sticking problem thoroughly. On the other hand, {\bos} can solve this problem by creating different training sets for $g$ input models, which means to let models converge to different local optima. So there will be more gaps and divergence among the learned parent models, which can potentially enhance the advantages of the {\ga} and base learners to achieve better solutions.

\subsubsection{Boosting Strategy}

The overall framework of {\bos} based {\adam} is shown in Figure~\ref{fig:bos}. To explicitly explain the {\bos} strategy, first, we have to introduce the weight $\mb{z} \in \mathbb{R}^m$ for all the training instances (here, $m$ denotes the size of the total training set). For each input model's training, one sub training set is sampled subject to the current weight vector $\mb{z}$. In other words, the value of $\mb{z}[i]$ represents the probability that $i_{th}$ training instance will be sampled. Initially, $\mb{z}[i] = \frac{1}{m}, \forall i \in \{1, 2, \cdots, m \}$, i.e., all the instances will be sampled with an equivalent chance. We train the first input model $M_1$ by {\adam} with a sub training set sampled from the entire training set subject to the weight vector $\mb{z}$. After this training process, we can denote the trained first model as $\bar{M}_1$ and record its prediction results on the entire training set to update $\mb{z}$ by
\begin{equation}
\mb{z}[i] = \mb{z}[i]\times \left\{ \begin{array}{ll}
\exp(-\alpha_j) & \mbox{if $\bar{M}_j(\mb{X}[i,:]) = \mb{y}[i]$};\\
\exp(\alpha_j) & \mbox{if $\bar{M}_j(\mb{X}[i,:]) \not= \mb{y}[i]$}.\end{array} \right.
\label{equ:W1}
\end{equation}
and
\begin{equation}
\mb{z}[i] = \mb{z}[i]/ sum(\mb{z})
\label{equ:sum}
\end{equation}
where $\bar{M}_j(\mb{X}[i,:])$ denotes the output of model $\bar{M}_j$ on instance $\mb{X}[i,:]$, $\alpha_j = \frac{1}{2}log(\frac{1-\epsilon_j}{\epsilon_j})$ and $\mb{y}[i]$ represents the true label of $i_{th}$ training example. Here, for the first input model $\bar{M}_1$, $j = 1$ and $\epsilon_1 = P_{(\mb{X}[i,:], \mb{y}[i])\sim \mathcal{X}}(\bar{M}_1(\mb{X}(i,:)) \not= \mb{y}[i])$. What need to be mentioned is that $\epsilon_j <0.5$ should be satisfied because we deliberately design the weight of training sample $\mb{X}[i,:]$ having $\bar{M}_1(\mb{X}[i,:]) \not= \mb{y}[i]$ to increase, so $\alpha_1$ will be larger than 0. In this way, sample $\mb{X}[i,:]$ will be more likely to appear in next sub training set and the next model will focus more on correctly classifying it. The Equation (\ref{equ:sum}) is to regulate $\mb{z}$ as a probability distribution. When training the next model by {\adam}, another sub training set will be sampled based on the new weight vector $\mb{z}$. By using the weight vector $\mb{z}$, the {\bos} strategy successfully assigns entire training set with different distributions (redistribution of the training set). 

The {\bos} strategy can also be regarded as the interaction among models. By applying {\bos} we achieve the interaction among models through updating the sub training set for each input model. What is more, this interactive training essentially makes these $g$ input models diverse. Since different training sets will lead to various loss functions and gradients, $g$ input models are more likely to converge to other local optimal points. This characteristic can give {\ga} more advantages, and that is the core idea we implement the {\bos} strategy into the {\yb}.

%-------------------------------------
%--------------------------------------

%\subsubsection{Combining Boosting with {\adam}}

%From the above content we can notice that, {\bos} strategy and the input model's triaining are conducted interactively. 

\section{THEORETIC ANALYSIS}\label{sec:theorecital_analysis}

%Under the framework introduced in Section~\ref{sec:method}, {\yb} can not only guarantee the effectiveness of
In this section, we will show that the proposed {\yb} algorithm can guarantee the learning effectiveness; meanwhile, keep converging when the number of generations increases.

\subsection{Effectiveness Guarantee}
According to~\cite{boosting}, it has been proven that the error $\epsilon_i$ of the trained input model $\bar{M}_i$ can be bounded by
%\vspace{-5pt}
\begin{equation}
2^g \prod_{i = 1}^{g}\sqrt{\epsilon_i(1 - \epsilon_i)} \leq \exp(-2\sum_{i=1}^{g}\gamma_i^2)
\end{equation} 
where $\gamma_i = 0.5 - \epsilon_i$. Let $\epsilon_{min}$ be the error of input model with the smallest loss, we have 
%\vspace{-5pt}
\begin{equation}
\begin{aligned}
2^g \epsilon^g_{min}\leq 2^g (\epsilon_{min}(1-\epsilon_{min}))^{\frac{g}{2}} \leq 2^g \prod_{i = 1}^{g}\sqrt{\epsilon_i(1- \epsilon_i)}
\end{aligned}
\end{equation}
Therefore $\epsilon_{min}\leq \frac{1}{2}\exp(-2\sum_{i=1}^{g}\gamma_i^2/g)$. The error of the model we finally select as the output model in the last generation has a relatively tight upper bound.
%------------------------------------------------------
\begin{table*}[t]
	\vspace{-15pt}
	\renewcommand\arraystretch{1.2}
	\centering
	\begin{threeparttable}
		\caption{Experiment results on all datasets}
		\begin{tabular}{|c|c|c|c|c|c|c|}
			\hline 
			\multirow{3}*{\textbf{Comparison Methods}}&\multicolumn{6}{|c|}{Datasets}\\
			
			\cline{2-7}
			&\multicolumn{2}{|c|}{ORL-7\tnote{*}}&\multicolumn{2}{|c|}{MNIST}&\multicolumn{2}{|c|}{CIFAR-10}\\
			\cline{2-7}
			&Accuracy&Loss&Accuracy&Loss&Accuracy&Loss\\
			\hline
			{\yb}&0.967&\textbf{0.0947}&\textbf{0.9917}&\textbf{0.045}&\textbf{0.6358}&1.2346\\
			\hline
			{\yadam}&\textbf{0.975}&0.1658&0.9911&0.076&0.6189&2.4067\\ 
			\hline
			{\adam}&0.958&0.466&0.9905&0.0458&0.6103&\textbf{1.1742}\\
			\hline
			RMSProp&0.9417&0.2819&0.9877&0.0464&0.5978&1.1807\\
			\hline
			AdaGrad&0.9333&1.763&0.7988&1.5457&0.3292&1.6892\\
			\hline
		\end{tabular}
		\label{tab:l2}
		\begin{tablenotes}
			\footnotesize
			\item[*] ORL-7 denotes 3 images per person as test set, the rest 7 images as training set and 4 of 7 as validation set.
		\end{tablenotes}
	\end{threeparttable}
	\vspace{-10pt}
\end{table*}
%--------------------------------------------------------

\subsection{Convergence Analysis}\label{subsec:convergence_ana}
In our proposed method, we effectively integrate {\bos} based {\adam} learning algorithm into {\yb} as shown in Figure~\ref{fig:yb} and Algorithm~\ref{alg:gadam}. For each generation $G^{(k)}$, while training $g$ input models, we no longer use the same training set for every model; instead, the {\bos} based learning algorithm shown in Figure~\ref{fig:bos} is added to the training process. For $M_i$, the training set $\mathcal{D}_i$ (denoted by a gray oval) is applied. A remark to be added here, the training process of the child models also applies the {\bos} strategy as indicated in the algorithm architecture. With the growth of the $k$, we will show that the loss of input models can finally converge.

%The convergence of {\yadam} has been proven in. Similar to the convergence analysis in~\cite{yadam},
The {\yb} algorithm will also converge in a finite number of generations, along with the loss of the input models decreasing continuously. For the $i_{th}$ input model $M_i^{(k)}$ in the generation $G^{(k)}$, the variables learned by {\bos} based {\adam} will converge after training, which means $\bar{\mb{loss}}_i^{(k)} \leq \mb{loss}_i^{(k)}$, where $\mb{loss}_i^{(k)}$ and $\bar{\mb{loss}}_i^{(k)}$ denote the introduced loss of the model $M_i$ and $\bar{M}_i$ before and after training respectively in generation $G^{(k)}$. Since in the $k_{th}$ generation,  the top $g$ models are selected among both the trained input models and the generated child models as the output models, the performance of those chosen $g$ models will not be worse than the input models. In other words, $\mb{loss}_i^{(k+1)} \leq\bar{\mb{loss}}_i^{(k)}, \forall i \in \{1,2,\dots, g\}$. Therefore we have 
%\begin{small}
\begin{equation}
\begin{split}
%\vspace{-5pt}
\mb{loss}^{(k+1)} = \sum_{i=1}^{g} \mb{loss}_i^{(k+1)}
\leq \sum_{i=1}^{g}\bar{\mb{loss}}_i^{(k)} \leq \sum_{i=1}^{g}\mb{loss}_i^{(k)}
=\mb{loss}^{(k)}
\end{split}
\end{equation}
%\end{small}
With the training process going on, the loss of models in each generation will continuously decrease when $K$ goes up, and the {\yb} will finally converge. The convergence results will be exhibited in Section~\ref{subsec: alg_convergence}.

%------------------------------------------
\section{NUMERICAL EXPERIMENTS}\label{sec:experiment}
\vspace{-5pt}
To test the effectiveness and advantages of the proposed {\yb} algorithm, extensive experiments have been conducted on real-world datasets. In this section, we will first describe the datasets we have used in the experiment, and then introduce the experimental settings in detail. Finally, we will exhibit the experimental results together with detailed descriptions and give the parameter sensitivity analysis.

\subsection{Dataset Description}
\begin{itemize}
	\item \textbf{ORL Dataset}: The ORL~\cite{ORL} dataset consists of face images of 40 people, each person has ten images. Each image is in size of 112$\times$92.
	\item \textbf{MNIST Dataset}: The MNIST~\cite{GBDR} dataset includes 60,000 training samples and 10,000 testing samples, where each sample is a 28$\times$28 image of hand-written numbers from 0 to 9. 
	\item \textbf{CIFAR-10 Dataset}: The CIFAR-10~\cite{CIFAR} dataset consists of 60000 32$\times$32 color images in 10 classes, with 6000 images per class. There are 50000 training images and 10000 test images. The dataset has no augmentation operation.
\end{itemize}
%These above datasets have different sizes, but all of them are image data. For ORL dataset, it only has 400 samples in total; for MNIST dataset, the scale is much larger. Next we will also discuss the data implementation details with experiment settings. 
\vspace{-5pt}
\subsection{Experiment Settings}
In this part, we will introduce the experiment settings, which covers the detailed experiment setup, comparison methods, and the evaluation metrics.
\subsubsection{Experiment Setup}

We use the convolutional neural network (CNN) structure models as the base model (input model). The CNN model we have built is based on the LeNet-5 \cite{GBDR}, which has seven layers. %(convolution layer 1, max pooling layer 1, convolution layer 2, max pooling layer 2, fully connected layer 1, fully connected layer 2, output layer). 
For different datasets, the CNN models have different settings:
%For MNIST dataset, we use LeNet-5 structure, while 
for the ORL dataset, we use two convolutional layers with 16 and 36 feature maps of $5\times 5$ kernels and $2\times 2$ max-pooling layers, and a fully connected layer with 1024 neurons; for the MNIST dataset, we use LeNet-5 structure in CNN model; for the CIFAR-10 dataset, we apply three convolutional layers with 64, 128, 256 kernels respectively, and a fully connected layer having 1024 neurons.
% for CIFAR-10 dataset, we use three convolutional layers with 64,128,256 kernels respectively, and a fully connected layer having 1024 nodes. 
All the experiments apply Relu \cite{relu} activation function, and 0.5 dropout rate on fully connected layers. 
%\item \textbf{Optimization algorithms}
For each training process, the training batch involves 128 samples and the number of epochs satisfies traversing the entire training set around 200 times. The initialization of variables is random numbers sampled from Normal Distribution with $0$ means and $0.01$ standard variance. 
%\item \textbf{{\yadam}}:  
%To show the advantages of adding {\bos} strategy, we implement {\yadam} algorithm with $K$ as the comparison method. 
%\item \textbf{{\yb}}: 
For different datasets we use different $g$ and $K$ in the {\yb}, and we will analyze their influence later.
%For ORL dataset, we set $K = 10$, $g = 10$ to get the output result.
%for MNIST dataset, $K=5$ and $g = 5$; for CIFAR-10 dataset,$K = 3$ and $g= 5$.

%\end{itemize}
%We set the learning epoch of {\adam} to 200, as we find the training accuracy and loss converge at this point. We will first apply the CNN model to the datasets mentioned above, whose performance can be measured as the ground truth.

\subsubsection{Comparison Methods and Evaluation Metrics}
To show the advantages of the {\yb} algorithm, we compare it with the most commonly used optimization algorithms, including {\adam}~\cite{ADAM}, RMSProp~\cite{rmsprop}, AdaGrad~\cite{adagrad} and {\yadam}~\cite{yadam} respectively.The  input model in our experiments is based on the CNN structures.

To measure the performance of the comparison methods, different metrics have been applied in this paper. We calculate both the accuracy of prediction and test loss achieved by the models trained with these different optimization algorithms on the test set.
%-------------------------------------------------------
\begin{figure}[t]
	\vspace{-10pt}
	\centering
	%	\hspace{-20pt}
	\subfigure[Loss on test set]{ \label{fig:g-K1}
		\begin{minipage}[l]{0.45\columnwidth}
			\centering
			\includegraphics[width=1\textwidth,height=2.9cm]{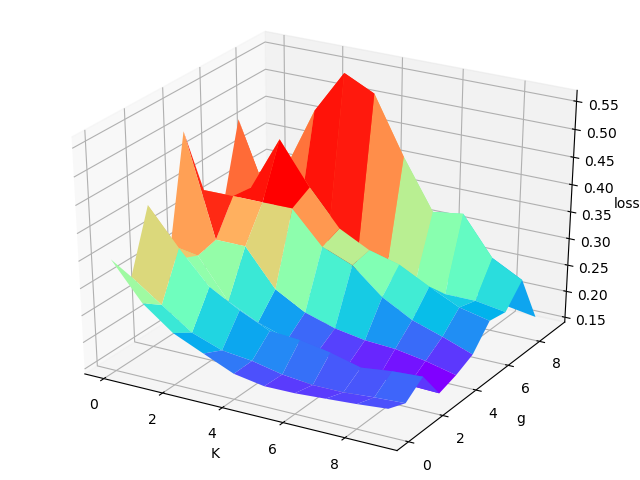}
		\end{minipage}
	}
	\hspace{-8pt}
	\subfigure[Accuracy on test set]{ \label{fig:g-K2}
		\begin{minipage}[l]{0.45\columnwidth}
			\centering
			\includegraphics[width=1\textwidth,height=2.9cm]{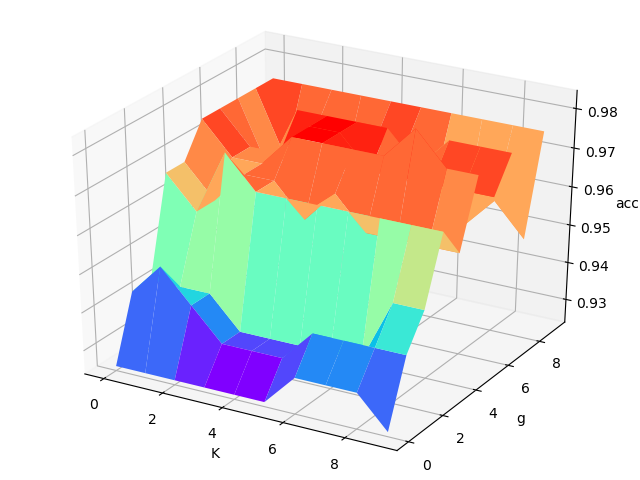}
		\end{minipage}
	}
	%\vspace{-8pt}
	\caption{The impact of $g$ and $K$}
	\label{fig:analysis}
	\vspace{-15pt}
\end{figure}
%--------------------------------------------------------
\vspace{-10pt}
\subsection{Experimental Results}\label{subsec: alg_convergence}
In this section, firstly, we will exhibit results on all the datasets, then show the convergence and hyper-parameter analysis results of the {\yb} algorithm.
%\vspace{-5pt}
\subsubsection{Results on All Datasets}
In Table~\ref{tab:l2}, we show the performance of {\yb} algorithm on several datasets compard with other baseline algorithms. The results of {\yb} in the table is with $g =5, K =5$. To let the total training iterations of {\yb} be identical to comparison methods (e.g., {\adam}), we set the training iterations in each generation of {\yb} as the training iterations of {\adam} dividing $K$. 
%In this way, the time-consuming of {\yb} is equal to the {\adam} if the $g$ models in each generation of {\yb} are run in parallel. 
In this way, the total number of training iterations of {\yb} is equal to {\adam}, meanwhile the overfitting problem can be potentially inhibited. From the table, we can see that the {\yb} achieves better overall performance on the test set. For the accuracy of prediction, {\yb} achieves the best results on most of the datasets, especially on the CIFAR-10 dataset. The test accuracy achieved by {\yb} is 0.6358, which is at least 2.5\% larger than the accuracy obtained by {\adam}, {\yadam} and {\rmsprop}, and the advantages are much more significant compared with {\adagrad}; for the loss on the test set, the advantages of {\yb} are much more obvious: the {\yb}'s results are less than {\yadam} and {\adam} by almost 50 percent on all datasets. Especially for the ORL and CIFAR-10 datasets, the test loss of {\yb} is only half of {\yadam}. The advantages are much more significant when comparing to other algorithms such as {\rmsprop} and {\adagrad}. The overall results demonstrate that our proposed {\yb} method does improve the performance of both {\ga} and {\adam} by adopting the {\bos} strategy.

%-------------------------------------
\begin{figure}[t]
	%	\vspace{-20pt}
	\centering
	\includegraphics[width=0.32\textwidth,height=4cm]{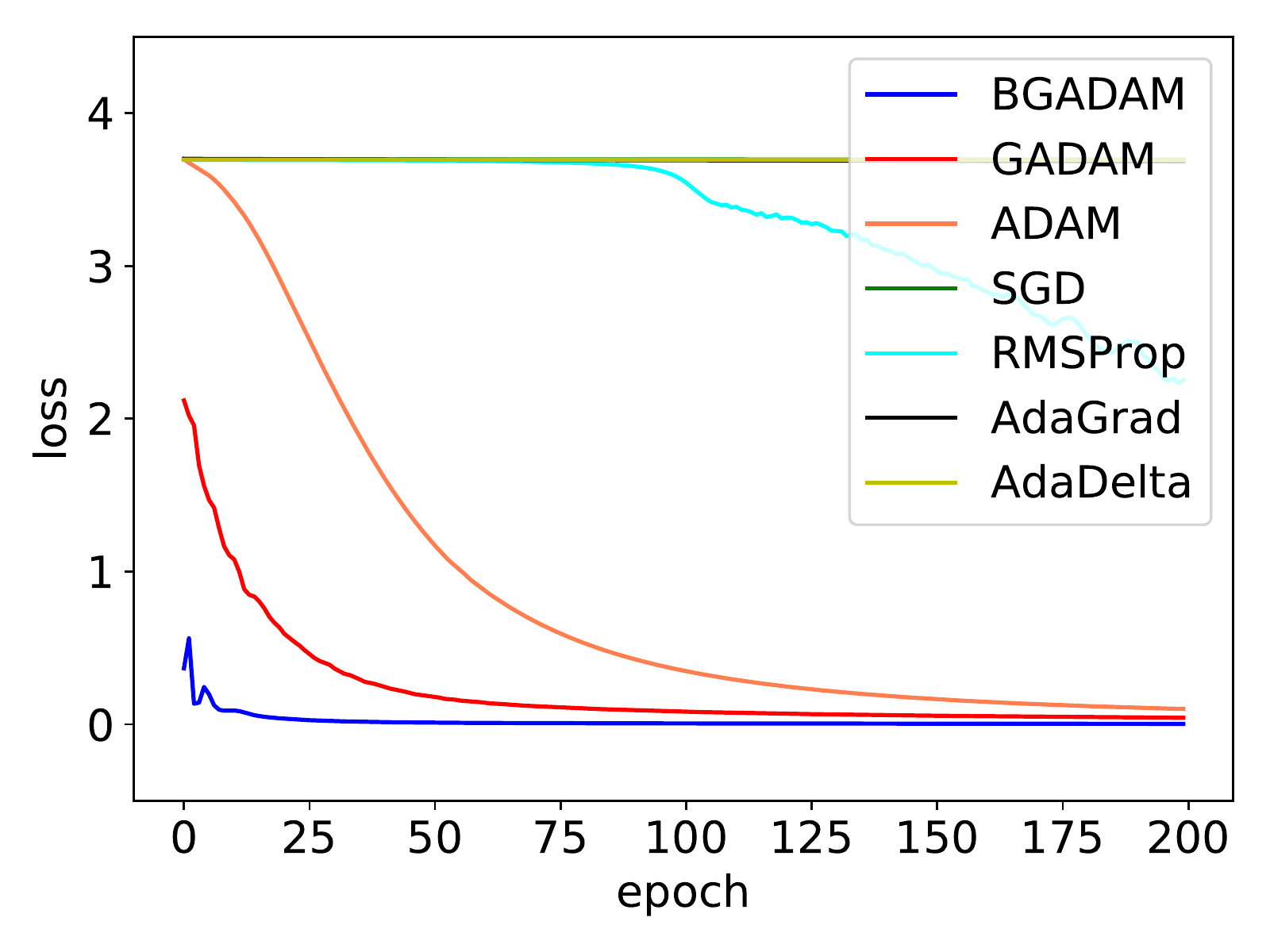}
	\vspace{-8pt}
	\caption{Convergence results of comparison algorithms}
	\label{fig:converge}
	\vspace{-10pt}
\end{figure}
%--------------------------------------
\subsubsection{Convergence and Hyper-Parameter Analysis}
{\yb} algorithm can converge in a finite number of generations, which is shown in Figure~\ref{fig:converge}. The result is on the ORL-7 dataset.
%From Figure~\ref{fig:converge} we can find that {\yb} converges faster than other comparison algorithms.
Due to the limited space, we will only show the analysis on the ORL dataset in this part. The hyper-parameters $g$ and $K$ involved in the training process may affect the convergence and final results of {\yb}. Thus we also analyze their influence. We run the experiments with $g \in \{1,2,\dots,10\}$ and $K\in \{1,2,\dots,10\}$ on the ORL dataset to illustrate their influence. The results are shown in Figure~\ref{fig:analysis}. 
We can notice that the test loss decreases when $K$ increases in terms of different $g$ numbers, and conversely, the accuracy on the test set increases. The trend can also be seen in Table~\ref{tab:l1}.
%-------------------------------------------------------
\begin{table}
	\vspace{-5pt}
	\renewcommand\arraystretch{1.1}
	\centering
	\caption{Hyper-parameter analysis}
	\begin{tabular}{|c|c|c|}
		\hline 
		\multirow{2}*{\textbf{Different Situations}}&\multicolumn{2}{|c|}{ORL-7}\\
		\cline{2-3}
		&Test loss&Test acc\\
		\hline
		$K = 1, g =10$&0.3935&0.975\\ 
		\hline
		$K=10,g=10$&0.088&0.9833\\
		\hline
	\end{tabular} 
	\label{tab:l1}
	\vspace{-10pt}
\end{table}
%--------------------------------------------------------
We also find that the loss function does not change when $K$ and $g$ increase from $5$ to $10$, respectively. In other words, the proposed {\yb} algorithm can converge to an ideal solution in $5$ generations with the input models size of $5$ generally. That is also why we apply the setting of $g = 5$ and $K = 5$ when carrying out the final results of {\yb}. For the experiments on the other datasets, we also check the influence of $K$ and verify specific values of $K$ to get the final model.
%\subsection{Algorithm Convergence}

%------------------------------------------
%\input{sec_ack.tex}
%------------------------------------------
\vspace{-6pt}
\section{CONCLUSION}\label{sec:conclusion}
\vspace{-7pt}
In this paper, we have introduced a new hybrid optimization algorithm, namely {\yb}. By combining {\adam}, {\ga}, and {\bos} strategy together, {\yb} can maximize the advantages of each part by utilizing their characteristics to efficiently jump out of local optima and prevent the ``local sticking'' phenomenon, then further converge to better solutions. We have carried out extensive experiments on real-world datasets, and the results show that our proposed {\yb} algorithm outperforms previous optimization methods, especially for training deep neural networks.
%------------------------------------------
\vspace{-5pt}
\section{Acknowledgement}\label{sec:ack}
\vspace{-5pt}
This work is partially supported by NSF through grant IIS-1763365.
%------------------------------------------

%\acks{Acknowledgements should go at the end, before appendices and references.}
%\newpage
\bibliographystyle{plain}
\bibliography{reference}

%\appendix
%
%\section{First Appendix}\label{apd:first}
%
%This is the first appendix.
%
%\section{Second Appendix}\label{apd:second}
%
%This is the second appendix.

\end{document}